\documentclass[11pt,a4paper]{article}
\usepackage[hyperref]{acl2018}
\usepackage{times}
\usepackage{latexsym}
\usepackage{url}
\usepackage{graphicx}
\usepackage{amsmath}
\usepackage{diagbox}
\usepackage{adjustbox}
\usepackage{multirow}
\usepackage{tabularx}

\aclfinalcopy % Uncomment this line for the final submission
\title{Recurrent Neural Networks with Pre-trained Language Model Embedding for Slot Filling Task}

\author{Liang Qiu, Yuanyi Ding, Lei He \\
University of California, Los Angeles, USA \\
  }

\date{February 22, 2018}
\begin{document}
\maketitle

\begin{abstract}
In recent years, Recurrent Neural Networks (RNNs) based models have been applied to the Slot Filling problem of Spoken Language Understanding and achieved the state-of-the-art performances. In this paper, we investigate the effect of incorporating pre-trained language models into RNN based Slot Filling models. Our evaluation on the Airline Travel Information System (ATIS) data corpus shows that we can significantly reduce the size of labeled training data and achieve the same level of Slot Filling performance by incorporating extra word embedding and language model embedding layers pre-trained on unlabeled corpora.
\end{abstract}

%\section{Credits}
%This paper has been inspired by the trail works in Semi-supervised sequence tagging with bidirectional language models by Peters, M. E., Ammar, W., Bhagavatula, C., Power, R. \shortcite{cite2}; The application of Recurrent Neural Network on slot filling by Mesnil, G., Dauphin, Y., Yao, K., Bengio, Y., Deng, L., Hakkani-Tur, D., ..., Zweig, G. \shortcite{cite1}; Bi-directional recurrent neural network by Vu, T. N., Gupta, P., Adel, H., and Schutze, H. \shortcite{cite5}; Multi-domain joint semantic frame parsing by Hakkani-Tür, D., Tür, G., Celikyilmaz, A., Chen, Y. N., Gao, J., Deng, L., Wang, Y. Y. \shortcite{cite7}. We also benefit a lot from Professor Lei He at UCLA ECE department and Professor Songchun Zhu at UCLA CS department.

\section{Introduction}
The Slot Filling task is a subtask of Spoken Language Understanding (SLU) and can be treated as a standard sequence labeling or sequence discrimination task \cite{cite6}. Figure 1 shows a typical sentence in the Airline Travel Information System (ATIS) dataset \cite{cite4} and its annotation of domain, intent, named entity and slot. Typically, the SLU will firstly recognize the sentence domain and intent. Then relying on a Slot Filling module, it extracts additional essential information to determine the appropriate response to users. 

The annotation of slots and named entities follows the IOB (Inside/Outside/Beginning) convention. The B- prefix before a tag indicates that the tag is the beginning of a chunk. An I- prefix before a tag indicates that the tag is inside a chunk. And an O tag indicates that a token belongs to no chunk. We can see that Slot Filling is similar to the Named Entity Recognition (NER) task,  while the slots are more specific than named entities. For example, the slot tag of "Boston" is B-departure while the named entity tag of it is B-city.
\begin{figure}[h!]
  \centering
  \includegraphics[width=0.46\textwidth]{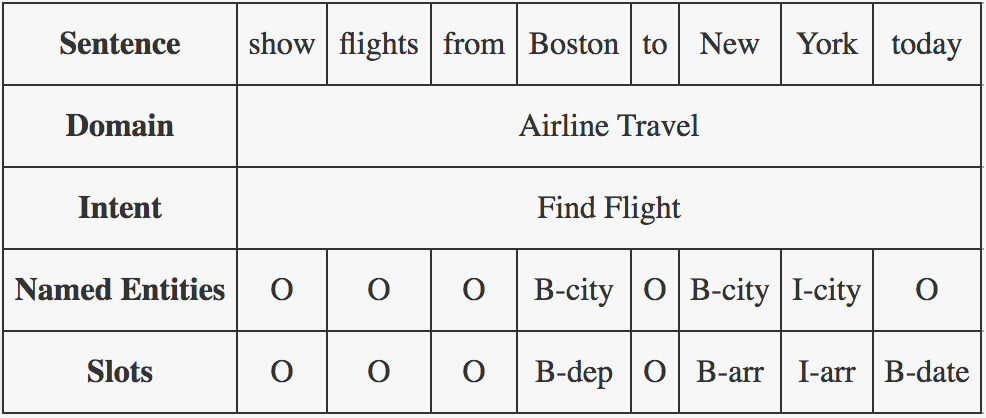}
  \caption{ATIS Utterance Example with the IOB Representation \cite{cite4}.}
\end{figure}

In recent years, Recurrent Neural Networks (RNNs) based models have been applied to the Slot Filling problem and achieved the state-of-the-art performances \cite{cite6}. However, training typical RNN models is often data-demanding, which limits its practical use in many specific domains where large amount of labeled training data is not available.  

In this paper, we investigate the effect of incorporating pre-trained word embedding and language models into RNN based Slot Filling models. Our evaluation on the Airline Travel Information System (ATIS) data corpus shows that incorporating an extra language model embedding layer pre-trained on an unlabeled corpus can significantly reduce the size of labeled training data without sacrificing the Slot Filling performance. Our results also suggest that using the pre-trained GloVe word embedding model and the bidirectional Long-Short Term Memory (bi-LSTM) model can achieve a better performance for the Slot Filling task.

\section{Related Work}
Modern methods to solve the Slot Filling problem include generative models such as Hidden Markov Model (HMM) \cite{cite8} and discriminative models such as Conditional Random Field (CRF) \cite{cite9}. With the popularity of RNNs in many other natural language processing (NLP) tasks such as language modeling \cite{cite10} and machine translation \cite{cite11}, RNN has also been applied to Slot Filling and achieved the state-of-the-art performance \cite{cite1}.

However, RNN models usually need to be trained with a large amount of labeled data to achieve the expected performance. The work presented in this paper has been inspired by previous work of using pre-trained fine-tuning word embedding models to improve the performance of deep learning based models (e.g., \cite{cite6}), and by the work of Peters et al. \shortcite{cite2} which used a pre-trained language model to encode the surrounding context of each word and improved the NER task performance. 

% Basically, a language model assigns a probability to the whole sequence:
% \begin{align*}
% 	p(t_1, t_2,..., t_N) &= \prod_{k=1}^N p(t_k | t_1, t_2,..., t_{k-1}) 
% \end{align*}

%We also use word embedding, where words or phrases from the vocabulary are mapped to vectors of real numbers. The word embedding can learn semantic and syntactic information of the words, i.e. similar words are close to each other in this space and dissimilar words far apart. These can be learned either using large amount of text like Wikipedia or specifically for a given problem. And previous work \cite{cite6} confirms that with pre-trained fine-tuning word vectors from SENNA brings out better performance. In this paper, we follow the ideas of RNN models combining with word embeddings, and further explore the sequence label performance of bidirectional LSTM and language models. 

\section{RNN with Language Model Embedding}
\subsection*{Overview}
In light of the success of RNNs in language modeling and many other natural language processing tasks, RNNs were introduced to solve slot filling problem which unsurprisingly achieved the state-of-the-art performance \cite{cite1}. But every coin has two sides. RNNs are usually data thirsty which means they need to be trained with a large amount of data to achieve the expected performance. In this section we discuss how to alleviate this shortcoming of RNNs with pre-trained language model embedding.

\subsection*{Baseline LSTM Model}
Our baseline model has basically the same structure as described in Mesnil et al. \shortcite{cite1} paper. But we replaced the simple Jordan and Elman versions of RNN with a modern two-layer LSTM.
\begin{align*}
    f_k &= \sigma(W_f \cdot [h_{k-1}, x_k] + b_f) \\
    i_k &= \sigma(W_i \cdot [h_{k-1}, x_k] + b_i) \\
    o_k &= \sigma(W_o \cdot [h_{k-1}, x_k] + b_o) \\
    \widetilde{C_k} &= \tanh(W_c \cdot [h_{k-1}, x_k]  + b_c) \\
    C_k &= f_k * C_{k-1} + i_k  * \widetilde{C_k} \\
    h_k &= o_k * \tanh(C_k)
\end{align*}
We will briefly refer LSTM as $ h_k = H(h_{k-1}, x_k) $ in the subsequent content. Our baseline LSTM model can be described as below.
\begin{align*}
	x_k &= w_k = E(t_k) \\
    h_k &= H(h_{k-1}, x_k) \\
    y_k &= softmax(W_y \cdot h_k + b_y) 
\end{align*}
where $ E() $ represents word embedding.

\subsection*{LSTM Model with Pre-trained Language Model Embedding}
Nowadays, using pre-trained word embedding such as Word2Vec or GloVe is quite popular in some NLP tasks. Before training the word embedding matrix, instead of initializing it with random values, initializing with pre-trained embedding can provide useful semantic and syntactic knowledge learned from another large dataset. However, for slot filling task, in addition to the meaning of a word, it's also important to represent the word in context. The state-of-the-art method (our baseline model) relies on the RNN model to encode the word sequences into a context-sensitive representation, which requires additional labeled data \cite{cite2}. Inspired by this work, we implemented a LSTM RNN model with language model embeddings pre-trained on the One Billion Word Benchmark \cite{cite3}, which contains one billion words and a vocabulary of about 800K words. As illustrated in Figure 2, the input sentence will be fed into both the GloVe and the pre-trained language model, and the result embeddings will then be concatenated as the new input embedding for the downstream 2-layer LSTM model.
\begin{align*}
	w_k &= E(t_k) \\
    l_k &= LM(t_1, t_2, t_3, ..., t_k) \\
    x_k &= [w_k, l_k]  \\
	h_k &= H(h_{k-1}, x_k) \\
    y_k &= softmax(W_y \cdot h_k + b_y)
\end{align*}
\begin{figure}[h!]
  \centering
  \includegraphics[width=0.49\textwidth]{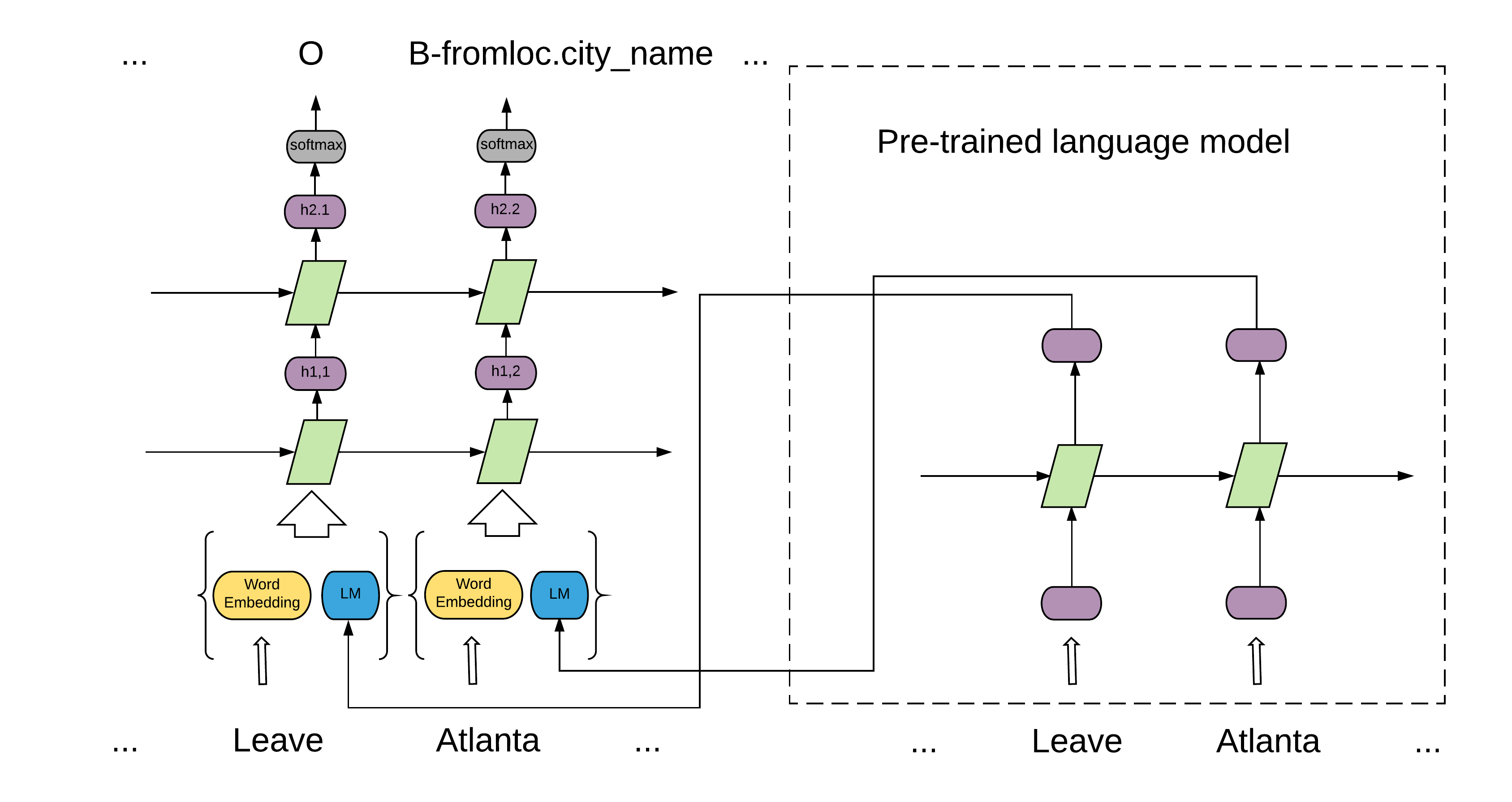}
  \caption{LSTM Model with Pre-trained Language Model Embedding.}
\end{figure}

\subsection*{Other Models Implemented}
Besides the models we talked above, we also implemented bidirectional LSTM with GloVe word embedding which is described as below:  

\begin{align*}
	w_k &= GloVe(t_k) \\
    l_k &= LM(t_1, t_2, t_3, ..., t_k) \\
    x_k &= [w_k, l_k] \\
	\overrightarrow{h_k} &= \overrightarrow{H}(\overrightarrow{h_{k-1}}, x_k) \\
    \overleftarrow{h_k} &= \overleftarrow{H}(\overleftarrow{h_{k+1}}, x_k) \\
    y_k &= softmax(\overrightarrow{W_y} \cdot \overrightarrow{h_k} + \overleftarrow{W_y} \cdot \overleftarrow{h_k} + b_y)
\end{align*}

We choose to use bi-directional recurrent neural network (bi-RNN) because combining information of the succeeding words is proved to be important for slot filling task \cite{cite5}. In bi-directional RNNs, words from both previous and future time step are regarded to predict the semantic tag of the target word. And GloVe can also provide a lot of useful addition semantic and syntactic information. We experimented all the combinations of these settings and in next section will talk about our experiment results.

\section{Evaluation}
\subsection*{Data}
We evaluate our model on the widely used Airline Travel Information System (ATIS) dataset \cite{cite4}. Words in the dataset are all labeled in the IOB format as shown in Table 1:
% \begin{figure}[ht!]
%   \centering
%   \includegraphics[width=0.46\textwidth]{atis.png}
%   \caption{Labeled data example in IOB format, where O denotes an artificial class for any words that do not have real slots}
% \end{figure}

\begin{table}[h]
\begin{tabular}{|p{2.2cm}<{\centering}|p{4.5cm}<{\centering}|}
\hline
Word & Label \\
\hline
what & O \\
flights & O \\
leave & O \\
% Atlanta & B-fromloc.city\_name \\
at & O \\
about & B-depart\_time.time\_relative \\
DIGIT & B-depart\_time.time \\
in & O \\
the & O \\
afternoon & B-depart\_time.period\_of\_day \\
and & O \\
arrive & O \\
in & O \\
San & B-toloc.city\_name \\
Francisco & I-toloc.city\_name \\
\hline
\end{tabular}
\caption{Labeled Sentence Example in IOB Format.}
\end{table}

\subsection*{Models}
Using Tensorflow, We first implement a two-layer LSTM network as our baseline model. Then we compare it with the proposed LSTM with language model embedding and another two complexer architectures, and report F1 scores on the testing dataset for each of the followings:
\begin{itemize}
\item \textbf{Baseline LSTM:} A forward RNN with LSTM cells, where the language model embedding and GloVe word embedding are not included.
\end{itemize}
\begin{itemize}
\item \textbf{LSTM + LM:} A forward LSTM (baseline) with pre-trained language model embedding, where the GloVe word embedding is not included.
\end{itemize}
\begin{itemize}
\item \textbf{Bi-LSTM + LM:} A bi-directional LSTM with pre-trained language model embedding, where the GloVe word embedding is not included.
\end{itemize}
\begin{itemize}
\item \textbf{Bi-LSTM + LM + Glove}: A bi-directional LSTM with pre-trained language model embedding and GloVe word embedding.
\end{itemize}
For the model architectures with GloVe word embedding, we assign the pre-trained embedding vectors to corresponding words in our dataset that can be found in the GloVe dictionary and randomly initiate the embeddings of the rest words. For models without GloVe embedding, we just randomly initiate the whole word embedding matrix. We use the RMSPropOptimizer with a constant learning rate $ \alpha = 0.001 $ to reduce the cross entropy loss. We also adopt dropout with 80\% keep probability to avoid over fitting.

\subsection*{Results}
We train the four models with variable dataset size from 100 sentences to 3600 sentences. Then we track the loss function value during the training process and find that 40 training epochs are sufficient for the optimizer to converge. So we test the performance of the four different models on a test dataset of 893 sentences after 40 epochs and plot their F1 scores in Figure 3. 

Observing the trend of each curve, we see that the F1 score increases rapidly with the increasing dataset size while the number of sentences is under 600. But keeping increasing dataset size above that number does not help improve the performance much since the amount of data is already sufficient to train the models with our defined complexity. 

\begin{figure}[ht!]
  \centering
  \includegraphics[scale=0.072]{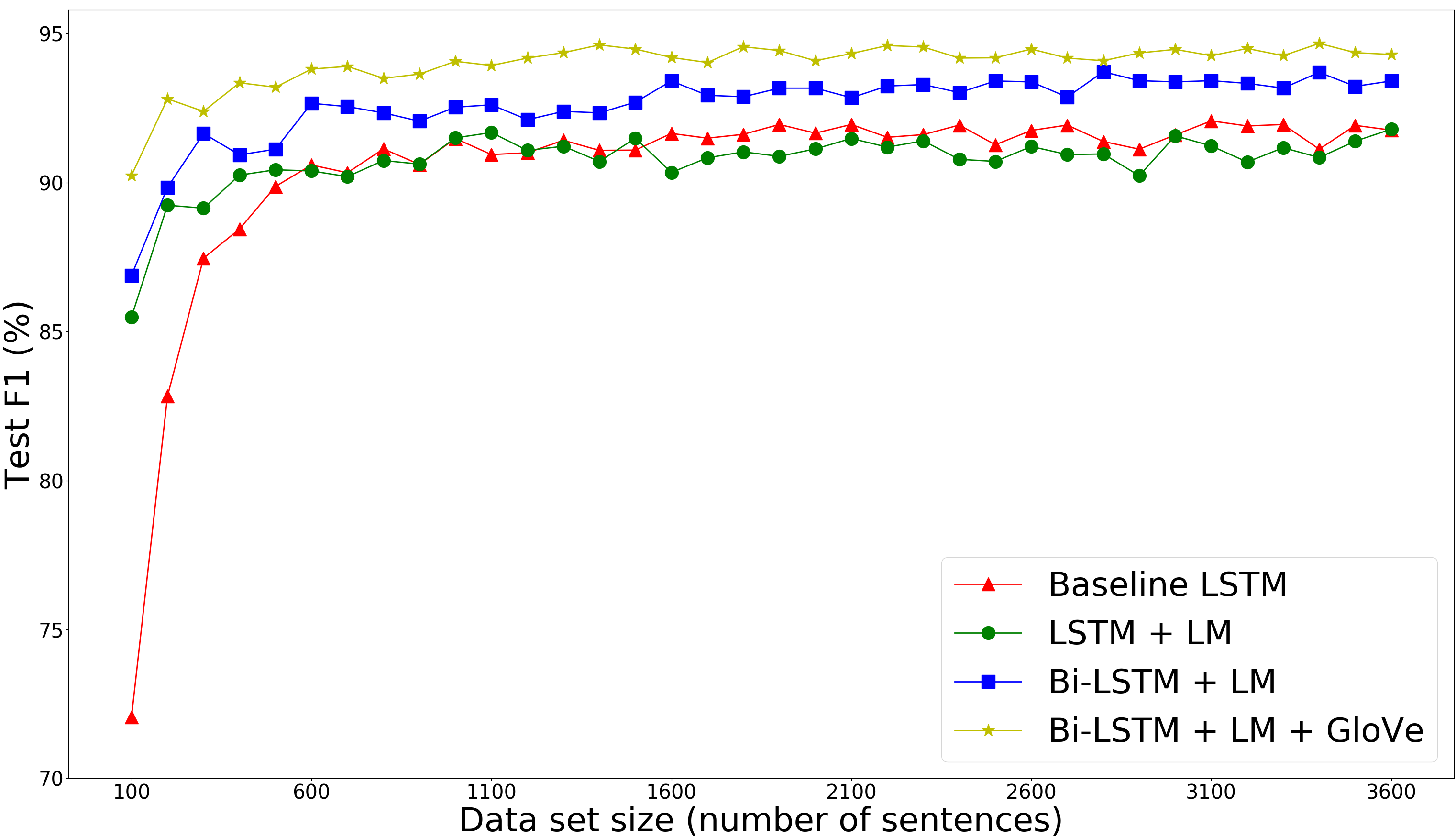}
  \caption{F1 Scores of Different Model Architectures Trained with Variable Dataset Sizes.}
\end{figure}

\begin{table}[h]
\footnotesize
\begin{tabular}{ |p{1.3cm}<{\centering}|p{3.8cm}<{\centering}|p{1.2cm}<{\centering}| }
\hline
Data Size & Model & F1 Score \\
\hline
\multirow{4}{3em}{100} & Baseline LSTM & 72.06\\
& LSTM + LM & 85.49 \\
& Bi-LSTM + LM & 86.88 \\
& Bi-LSTM + LM + GloVe & 90.24 \\
\hline
\multirow{4}{3em}{200} & Baseline LSTM & 82.83\\
& LSTM + LM & 89.24 \\
& Bi-LSTM + LM & 89.84 \\
& Bi-LSTM + LM + GloVe & 92.81 \\
\hline
\multirow{4}{3em}{300} & Baseline LSTM & 87.46\\
& LSTM + LM & 89.14 \\
& Bi-LSTM + LM & 91.65 \\
& Bi-LSTM + LM + GloVe & 92.39 \\
\hline
\multirow{4}{3em}{400} & Baseline LSTM & 88.44\\
& LSTM + LM & 90.25 \\
& Bi-LSTM + LM & 90.93 \\
& Bi-LSTM + LM + GloVe & 93.35 \\
\hline
\multirow{4}{3em}{500} & Baseline LSTM & 89.87\\
& LSTM + LM & 90.43 \\
& Bi-LSTM + LM & 91.12 \\
& Bi-LSTM + LM + GloVe & 93.2 \\
\hline
\multirow{4}{3em}{1000} & Baseline LSTM & 91.48\\
& LSTM + LM & 91.5 \\
& Bi-LSTM + LM & 92.53 \\
& Bi-LSTM + LM + GloVe & 94.07 \\
\hline
\multirow{4}{3em}{1500} & Baseline LSTM & 91.09\\
& LSTM + LM & 91.49 \\
& Bi-LSTM + LM & 92.7 \\
& Bi-LSTM + LM + GloVe & 94.48 \\
\hline\multirow{4}{3em}{2000} & Baseline LSTM & 91.66\\
& LSTM + LM & 91.13 \\
& Bi-LSTM + LM & 93.17 \\
& Bi-LSTM + LM + GloVe & 94.09 \\
\hline\multirow{4}{3em}{2500} & Baseline LSTM & 91.27\\
& LSTM + LM & 90.71 \\
& Bi-LSTM + LM & 93.41 \\
& Bi-LSTM + LM + GloVe & 94.19 \\
\hline\multirow{4}{3em}{3000} & Baseline LSTM & 91.6\\
& LSTM + LM & 91.57 \\
& Bi-LSTM + LM & 93.38 \\
& Bi-LSTM + LM + GloVe & 94.47 \\
\hline
\end{tabular}
\caption{Part of the F1 Scores in Figure 3.}
\end{table}

For detailed results and comparison, we also list the F1 score values with respect to different training data sizes in Table 2.
By comparing the F1 scores of different models, we find that adding pre-trained language model embedding can significantly improve the performance of LSTM, especially when the training dataset is relatively small. For instance, with only 100 and 200 training examples, our best model (Bi-LSTM+LM+Glove) outperforms the baseline LSTM model by large margins of 18\% and 10\% respectively. Such results demonstrate the great potential of our Bi-LSTM+LM+Glove based Slot Filling to be used in practical domains where labeled training data are difficult/expensive to get.

As the training data size increases, the benefit of incorporating pre-trained language model embedding becomes less significant since the training dataset is large enough for the baseline LSTM to learn a good context model. Nevertheless, we can still conclude that besides the proposed language model embedding, GloVe word embedding and bi-directional LSTM help to improve the model performance for the slot filling task as well. With sufficient amount of labeled training data ($>1000$), the Bi-LSTM+LM+Glove still outperforms the baseline LSTM model by nearly 3\%.

% \begin{figure}[ht!]
%   \centering
%   \includegraphics[scale=0.072]{valid_f1.png}
%   \caption{F1 score comparison between different model architectures trained with different data sizes}
% \end{figure}

% \begin{table}[h]
% \footnotesize
% \begin{tabular}{ |p{1.3cm}|p{3.8cm}|p{1.2cm}| }
% \hline
% Data Size & Model & F1 Score \\
% \hline
% \multirow{4}{3em}{500} & Baseline LSTM & 94.4\\
% & LSTM + LM & 95.7 \\
% & Bi-LSTM + LM & 96.35 \\
% & Bi-LSTM + LM + GloVe & 97.02 \\
% \hline
% \multirow{4}{3em}{1000} & Baseline LSTM & 95.28\\
% & LSTM + LM & 96.03 \\
% & Bi-LSTM + LM & 97.44 \\
% & Bi-LSTM + LM + GloVe & 97.65 \\
% \hline
% \multirow{4}{3em}{1500} & Baseline LSTM & 95.73\\
% & LSTM + LM & 96.28 \\
% & Bi-LSTM + LM & 97.65 \\
% & Bi-LSTM + LM + GloVe & 98.01 \\
% \hline\multirow{4}{3em}{2000} & Baseline LSTM & 96.23\\
% & LSTM + LM & 96.25 \\
% & Bi-LSTM + LM & 97.45 \\
% & Bi-LSTM + LM + GloVe & 98.03 \\
% \hline\multirow{4}{3em}{2500} & Baseline LSTM & 96.1\\
% & LSTM + LM & 96.51 \\
% & Bi-LSTM + LM & 97.76 \\
% & Bi-LSTM + LM + GloVe & 98.08 \\
% \hline\multirow{4}{3em}{3000} & Baseline LSTM & 96.23\\
% & LSTM + LM & 96.52 \\
% & Bi-LSTM + LM & 97.7 \\
% & Bi-LSTM + LM + GloVe & 98.18 \\
% \hline
% \end{tabular}
% \caption{Part of the F1 scores in Figure 4}
% \end{table}

\section{Conclusion and Future Work}
\label{sec:length}
In this paper, we proposed a bi-directional LSTM model with pre-trained language model embedding and GloVe word embedding for slot filling task. This model significantly improves the recognition performance compared to the baseline LSTM model, especially under the situation where we do not have enough labeled sentences in a specific domain. 

One envisioned future work is to explore what other kind of general knowledge can be learned from public resources and embedded into the model for a specific task domain.

\bibliography{acl2018}
\bibliographystyle{acl_natbib}
\end{document}